\documentclass[sigconf]{acmart}
\usepackage{amsmath}
\usepackage{multirow}
\usepackage{subcaption}
\usepackage{import}
\usepackage{textcomp}
\usepackage{tikz}
\usetikzlibrary{positioning}

\DeclareMathOperator*{\argmax}{arg\,max}

\AtBeginDocument{%
  \providecommand\BibTeX{{%
    \normalfont B\kern-0.5em{\scshape i\kern-0.25em b}\kern-0.8em\TeX}}}

\acmConference[Woodstock '18]{Woodstock '18: ACM Symposium on Neural
 Gaze Detection}{June 03--05, 2018}{Woodstock, NY}
\acmBooktitle{Woodstock '18: ACM Symposium on Neural Gaze Detection,
 June 03--05, 2018, Woodstock, NY}
\acmPrice{15.00}
\acmISBN{978-1-4503-XXXX-X/18/06}

\begin{document}

\title{Illuminating Diverse Neural Cellular Automata for Level Generation}


\author{Sam Earle}
\email{sam.earle@nyu.edu}
\affiliation{%
  \institution{New York University}
  \state{New York}
  \country{USA}
}

\author{Justin Snider}
\email{justinsnider@nyu.edu}
\affiliation{%
  \institution{New York University}
  \state{New York}
  \country{USA}
}

\author{Matthew C. Fontaine}
\email{mfontain@usc.edu}
\affiliation{%
  \institution{University of Southern California}
  \state{California}
  \country{USA}
}

\author{Stefanos Nikolaidis}
\email{nikolaid@usc.edu}
\affiliation{%
  \institution{University of Southern California}
  \state{California}
  \country{USA}
}

\author{Julian Togelius}
\email{julian@togelius.com}
\affiliation{%
  \institution{New York University}
  \state{New York}
  \country{USA}
}








\renewcommand{\shortauthors}{Anonymous}

\begin{abstract}
  We present a method of generating diverse collections of neural cellular automata (NCA) to design video game levels. While NCAs have so far only been trained via supervised learning,
  we present a quality diversity (QD) approach to generating a collection of NCA level generators. By framing the problem as a QD problem, our approach can train diverse level generators, whose output levels vary based on aesthetic or functional criteria. To efficiently generate NCAs, we train generators via Covariance Matrix Adaptation MAP-Elites (CMA-ME), a quality diversity algorithm which specializes in continuous search spaces. We apply our new method to generate level generators
  for several 2D tile-based games: a maze game, \textit{Sokoban}, and \textit{Zelda}. Our results show that CMA-ME can generate small NCAs that are diverse yet capable, often satisfying complex solvability criteria for deterministic agents.
  We compare against a Compositional Pattern-Producing Network (CPPN) baseline trained to produce diverse collections of generators and show that the NCA representation yields a better exploration of level-space.
\end{abstract}



\maketitle

\section{Introduction}

We present a method for generating diverse collections of neural cellular automata (NCA) level generators.~\footnote{Code is available at \href{https://github.com/smearle/control-pcgrl}{https://github.com/smearle/control-pcgrl}}

NCAs were derived from cellular automata (CA) in artifical life research. CAs are inspired by morphogenesis, the process of forming complex multi-celled lifeforms originating from a single cell, through simple rules that govern changes in each cell's state by observing neighboring cell states. NCAs differ from CAs by representing these rules as a neural network which takes as input a grid of cell states and outputs the same grid of cells with a changed state. We draw attention to the similarity between the morphogenetic process and the incremental design process employed by game designers when authoring video game levels.

\begin{figure}[!t]
  \centering
  \includegraphics[width=\linewidth]{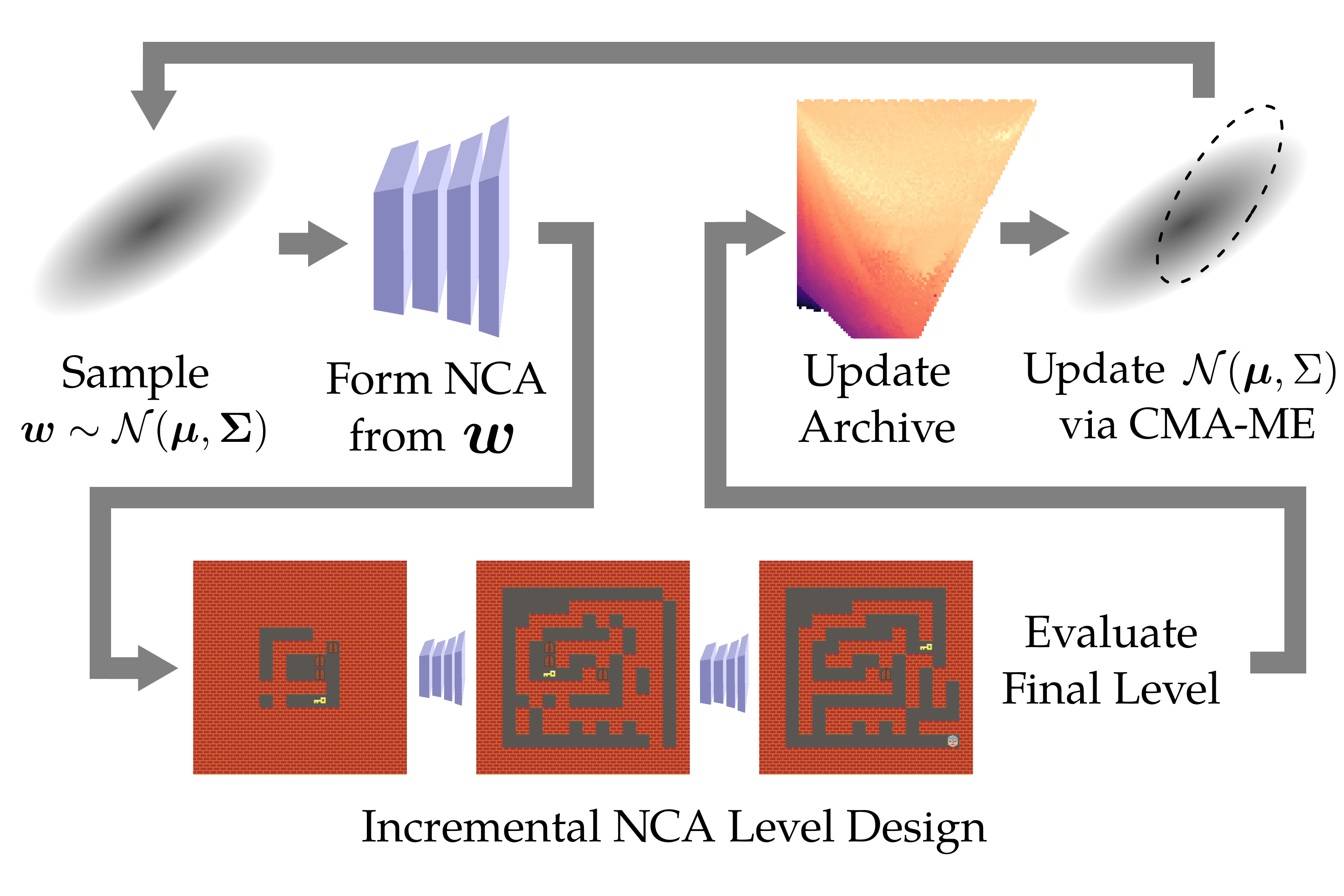}
  \caption{An overview of training NCAs for level generation. We sample NCA weights from a multi-variate Gaussian. The NCA repeatedly modifies levels for a fixed number of iterations. The final level is evaluated, then we add the NCA to the archive following CMA-ME's archive update.}
  \label{fig:front}
\end{figure}

NCAs could theoretically be used to simulate a wide variety of incremental processes involving local interactions---see, for example, the diversity of complex structures that have been discovered in both classic CAs ~\citep{cisneros2019evolving} and their variants~\citep{chan2020lenia, reinke2019intrinsically}.
But so far, NCAs have been
trained via supervised learning to learn a process that incrementally develops an initial state into a specific final artifact (e.g. replicating target images, textures, and 3D structures). However, we are interested in NCAs that generate levels satisfying specific criteria rather than a specific layout.
So rather than using NCAs as indirect encodings of target artifacts, we re-frame them as generators, that each produce a range of artifacts that all satisfy some global constraints.

To illustrate our method, let's consider a simple maze game where a player must walk from their initial position to an exit. Each maze must satisfy a criterion that all walkable areas must be connected. We can formulate the criterion as an objective of minimizing the number of connected regions present in the final output level after the NCA design process is complete. Formulating NCA training in the above way results in a single-objective optimization problem.

However, this objective can be satisfied by just two adjacent tiles containing the player and the exit with the remaining tiles filled with walls. To promote non-trivial solutions, we can also measure how far apart the player is from the exit or the number of empty tiles present in the final level. By including these measurable aspects as input to the optimization problem, we form a quality diversity optimization problem where we must find solutions that minimize the objective but are diverse with respect to the measurable criteria.

\begin{figure*}
  \centering
  \includegraphics[width=0.19\linewidth]{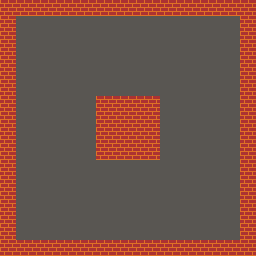}
  \includegraphics[width=0.19\linewidth]{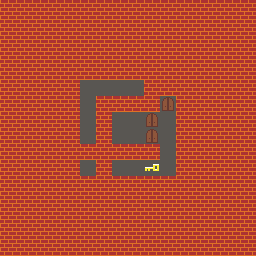}
  \includegraphics[width=0.19\linewidth]{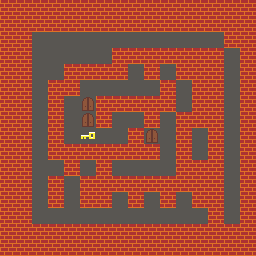}
  \includegraphics[width=0.19\linewidth]{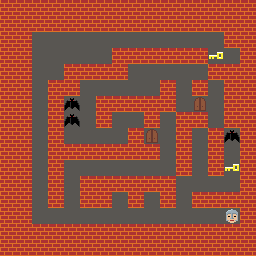}
  \includegraphics[width=0.19\linewidth]{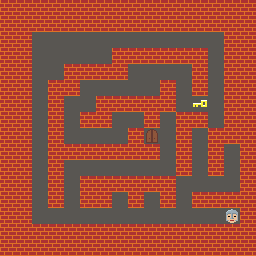}\\
  frame: 0 \hspace{0.18\linewidth}1 \hspace{0.18\linewidth}3 \hspace{0.18\linewidth}6 \hspace{0.18\linewidth}11
  \caption{\textbf{The NCA level generation process.} Information from a fixed initial seed is propagated out from the center of the board until a stable configuration, corresponding to high path-length, is reached. In intermediary level-states, tiles of different types, such as enemies and keys, appear to be used as a form of external memory during the generation process.}
  \label{fig:cellular_level_gen}
\end{figure*}

By framing the level-generation process as a QD problem, we thus generate a collection of NCAs, each producing levels of different length (or levels differing along any other, computable metric of form or function). But we may also want a variety of mazes with each length, and we may not be concerned about \textit{how} each maze achieves a given length. To encourage additional diversity beyond what is encoded as measured criteria, we treat each NCA like a generator, evaluating its performance on a set of random initial states -- here acting as latent inputs -- and adding to its objective a simple measure of diversity, i.e., the mean, pairwise per-tile difference over the set of generated levels.
This differs from previous approaches, which train NCAs on a single, fixed initial state. In an actual application, generating a set of level generators rather than just a set of levels has the considerable advantage that the trained level generator can output new levels many orders of magnitude faster than the search process which generates the generator~\cite{kerssemakers2012procedural}.



Our work makes the following contributions. 1) We formulate the training of an NCA as a quality diversity problem; 2) We train collections of level generators for several 2D tile-based games: a maze game, \textit{Sokoban}, and \textit{Zelda}; 3) We present qualitative results for each game and quantitative results based on the stability of the NCA generators for seeds outside the training seeds; and 4) We compare the use of incremental NCAs against a generative method using Compositional Pattern Producing Networks (CPPNs) to map $x, y$ coordinates and a latent $z$ to tile-values. To our knowledge, this is also the first work to apply NCAs to the problem of level-generation, and to use a derivative-free method for searching the space of NCA weights.

Overall, we reframe training NCAs from an incremental replication process to an incremental \textit{design} process, which designs novel video game levels that satisfy complex criteria specified by a human designer. Our proposed approach could eventually be used as a design aid to modify existing human-designed levels to satisfy human-specified criteria in a manner similar to~\citet{delarosa2021mixed}.

\section{Background}

Procedural Content Generation (PCG) can both render human designers more efficient, automating and abstracting away low-level tasks, and provide them with new tools for creative expression.
The sub-field of PCG via Machine Learning (PCGML~\citealt{summerville2018procedural}) studies training generative models on human-generated datasets, then prompting these models to generate plausible, original content at design-time.

One potential issue with PCGML is that models may be unlikely to extract functional constraints from data. One way of addressing this is to supplement the generative model with a specialized module tasked with satisfying functional constraints~\citep{zhang2020video}. Another is to forego data altogether, framing level design as a game in which the player is trained to meet these constraints~\cite{khalifa2020pcgrl}.
Our work takes the latter approach, but uses a specialized neural architecture to instantiate the generator.

Cellular Automata (CA) are systems involving localized, distributed computations, often over an unbounded 2D square grid.
Simple CAs have occasionally been used in PCG research: generating caves~\citep{johnson2010cellular}, influence maps~\citep{sweetser2005combining}, and maze-like levels~\citep{adams2017procedural}.
Many CAs are capable of producing complex, emergent dynamics over long time horizons~\citep{wolfram1984cellular, cisneros2019evolving}, making them a natural fit for content generation in spatial domains.

Neural Cellular Automata (NCAs,~\citealt{mordvintsev2020growing}) are CAs where rules are implemented as convolutional neural networks.
Conway's Game of Life, the canonical CA with complex dynamics, can be implemented in a very small NCA, though such an architecture is hard to train via gradient descent (unless over-parameterized)~\citep{springer2020s}.
Recently, more complex NCAs have been trained with gradient descent to generate images~\citep{mordvintsev2020growing}, textures~\citep{niklasson2021self}, and 3D structures~\citep{sudhakaran2021growing}.
These NCAs tend to be robust and capable of self-repair when faced with out-of-distribution perturbations.


In this work, NCAs perform an incremental level design task and are rewarded for the quality of the resultant levels according to a set of user-defined constraints and heuristics.
The NCA models replace the (at least $10\times$) larger neural networks that have previously been trained in this context using Reinforcement Learning (RL)~\cite{khalifa2020pcgrl}.

The end goal of our system is also similar to that of the controllable generators that have trained large networks using RL with conditional input and reward shaping~\citep{earle2021learning}.
Like the archives produced here, controllable generators are intended to be sampled by an end-user to produce levels with controllable features along measures of interest. However, our approach is to produce a diverse archive of specialists as opposed to a single jack-of-all-trades.


To optimize these smaller models, we turn to derivative-free quality diversity (QD) optimization methods~\citep{pugh2016quality, chatzilygeroudis2020quality}. Like RL, QD's problem definition allows for a variety of solving techniques. QD algorithms have been proposed based on genetic algorithms~\citep{mouret2015illuminating, cully:nature15, lehman2011evolving}, evolution strategies~\citep{conti2018improving, fontaine2020covariance, colas2020scaling}, differential evolution~\citep{choi:gecco21}, Bayesian optimization~\citep{kent2020bop}, and gradient ascent~\citep{fontaine2021differentiable}.

A quality diversity problem's definition includes an objective function to be maximized as well as a set of measure functions, which assign scalar values to each proposed solution. A QD algorithm will search for solutions which satisfy all output combinations of the measure functions and will break ties by maximizing the objective. We note that if measure functions are continuous, then an infinite number of solutions will exist. To resolve this issue, QD algorithms in the MAP-Elites~\citep{mouret2015illuminating, cully:nature15} family will pre-tesselate the space of the measure functions' outputs, which we refer to in this paper as a measurement space. The goal of the algorithm is then to find a solution to each cell of the tesselation, where ties are once again broken by the objective. In our NCA training problem, the objective and measure functions we propose are not easily differentiable. Therefore, we generate the collection of NCAs via a derivative-free QD algorithm based on evolution strategies.


Training the NCAs via evolutionary QD algorithms may yield additional benefits. Prior work demonstrates that evolutionary approaches are less prone to deception than their gradient descent counterparts~\citep{morse2016simple}. They may succeed in finding good solutions even when limited to small networks, while gradient descent would require much deeper neural networks to smooth the objective function~\citep{frankle2018lottery, nye2018efficient}.



\begin{figure}
  \centering
  \includegraphics[width=1.0\linewidth]{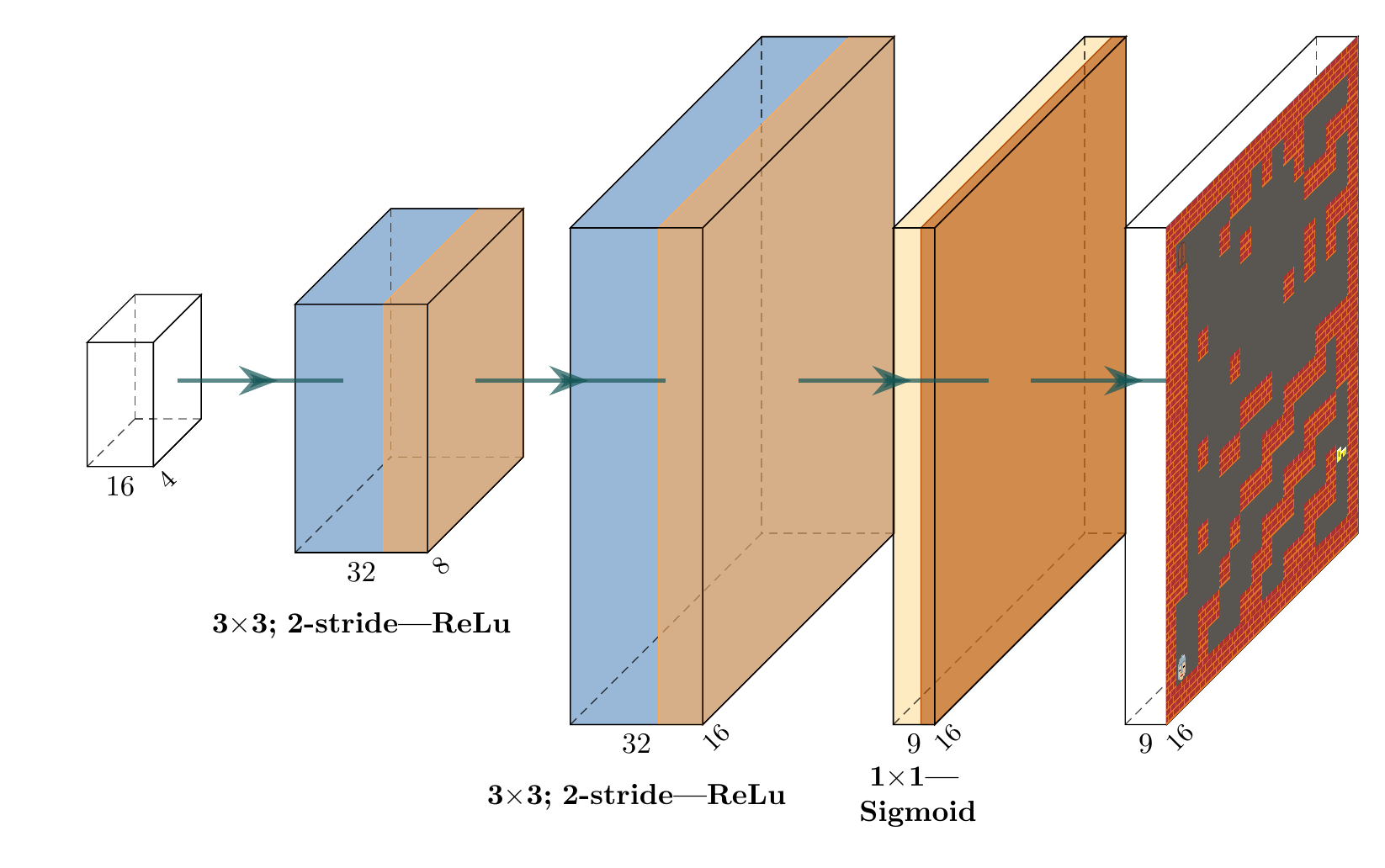}
  \caption{A decoder-like architecture maps random normal latent vector of size $16$ (tiled in 2D) to a game level in a single pass, with transposed strided convolutions. In the zelda domain, the architecture has $\approx 5,000$ parameters.}
  \label{fig:decoder_arch}
\end{figure}

\begin{figure}
  \centering
  \includegraphics[width=1.0\linewidth]{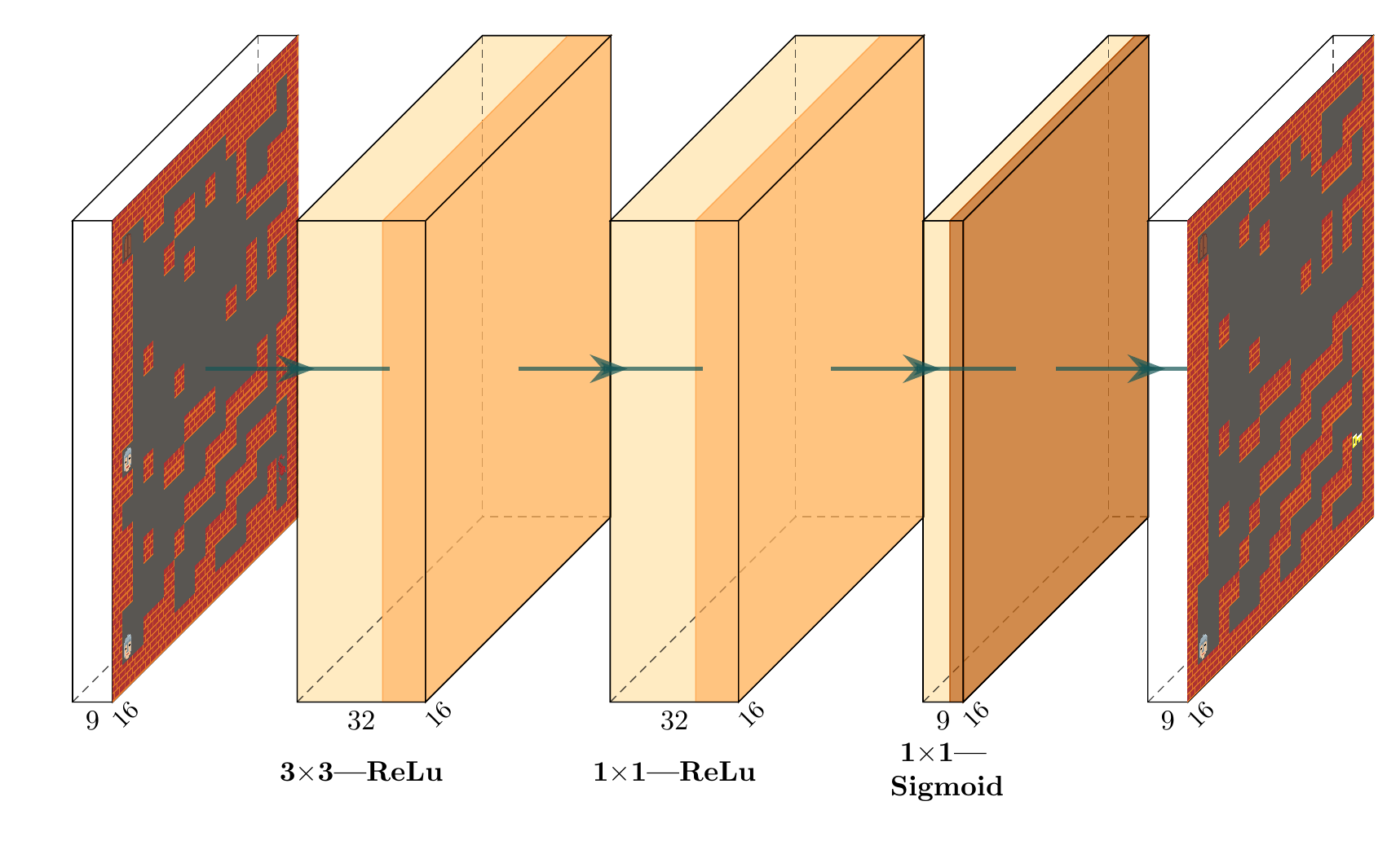}
  \caption{An NCA architecture repeatedly transforms a one-hot-encoded level state into another, based on local interactions over $3\times 3$ grids. In the zelda domain, the NCA has $\approx 3,000$ parameters.}
  \label{fig:nca_arch}
\end{figure}

\section{Methods}

We generate a diverse collection of NCA level-generators through \mbox{CMA-ME}~\citep{fontaine2020covariance}, a quality-diversity algorithm combining the adaptation mechanisms of \mbox{CMA-ES}~\citep{hansen2016cma,hansen2001completely} with the archiving
mechanisms of \mbox{MAP-Elites}~\citep{mouret2015illuminating}. We choose \mbox{CMA-ME} as it specializes in continuous domains and has been shown to be significantly more sample efficient than other QD algorithms in this setting. We train each NCA via the Pyribs library~\citep{pyribs:2021}, a QD optimization library maintained by the authors of \mbox{CMA-ME}.

The \textbf{NCA}s (Figure~\ref{fig:nca_arch}) have the same architecture as those of~\citet{springer2020s}, with 3 convolutional layers, ReLU and sigmoid activations, and $32$ hidden channels.
The output of the NCA has the same width, height, and number of channels as its input, with $\argmax$ applied channel-wise to produce a discrete encoding of the next level state. To have an NCA generate a game level, we feed it a onehot-encoded initial random level (or ``latent seed''), then repeat the process with the output of the NCA until the level converges to some stable state or a maximum number of steps is reached.

Once the model has generated a batch of levels, its objective value is computed along with any relevant measures, and this information is sent back to CMA-ME so that the algorithm may update the covariance matrix and add the model to the archive. Once we complete training, we evaluate the archive of NCA models on new latent seeds, observing the extent to which the models are capable of generalization.

A variation of the NCA architecture (\textbf{AuxNCA}) adds $3$ additional, ``auxiliary'' channels to the onehot-encoded level, which are initialized with 0s and, at subsequent iterations of the generative process, can serve as further external memory for the generative network. These channels are effectively invisible, and are separate from the onehot-encoding of the level.

We compare NCAs against Compositional Pattern-Producing Networks (\textbf{CPPN}s,~\citealt{stanley2007compositional}), which can generate spatial artifacts such as images~\citep{secretan2008picbreeder} by taking coordinate-pairs as input and outputting the multi-channel value at the corresponding pixel. CPPNs have previously been used to generate parts of game levels~\citep{team2021open}, and can naturally encode spatial patterns such as symmetry. We train CPPNs via MAP-Elites with the variation operations from the NEAT algorithm~\citep{stanley2002evolving}. MAP-Elites maintains an archive of diverse generators and mutates their architectures in search of new elites. We choose MAP-Elites over CMA-ME, since CMA-ME only works on fixed-length real-value vector representations and cannot easily be extended to augment network topology.

Additionally, we implement a fixed-topology CPPN with two hidden layers and sinusoidal activation functions (\textbf{SinCPPN}) and optimize its weights. We follow the weight-initialization scheme derived in~\citep{sitzmann2020implicit}, where it is shown that fixed-topology networks with periodic activations are capable of capturing fine-grained detail over continuous spaces. Having a fixed topology allows us to naturally make use of CMA-ME to train the CPPN's weights. 

It is important to note that since both CPPN-based models take only $(x, y)$ coordinates as input, they correspond to indirect encodings of individual levels rather than level-generators. To address this discrepancy, we implement generative versions of these models (\textbf{GenCPPN} and \textbf{GenSinCPPN}, inspired by~\cite{ha2016abstract}) that additionally take as input a random standard normal vector, which is tiled in 2D and concatenated to the $(x, y)$ coordinates at each tile before being fed into the CPPN.


Finally, we train networks comprising a series of transposed convolutions, mapping a small latent space to a one-hot level representation in a single pass (\textbf{Decoder}, Figure \ref{fig:decoder_arch}). These networks are inspired by the ``decoder'' subnetworks in Variational Auto-Encoders (VAEs) and in the generator networks of Generative Adversarial Network (GAN) training schemes for, e.g., image generation. As in the case of Generative CPPNs, the latent space involves a random standard normal vector, tiled in 2D and concatenated with $(x, y)$ coordinates, only this time over a smaller 2D grid (of $4\times 4$ tiles).

\begin{figure*}[t!]
  \begin{subfigure}[t]{0.49\textwidth}
    \begin{footnotesize}
      \centering
      \includegraphics[width=1.0\textwidth]{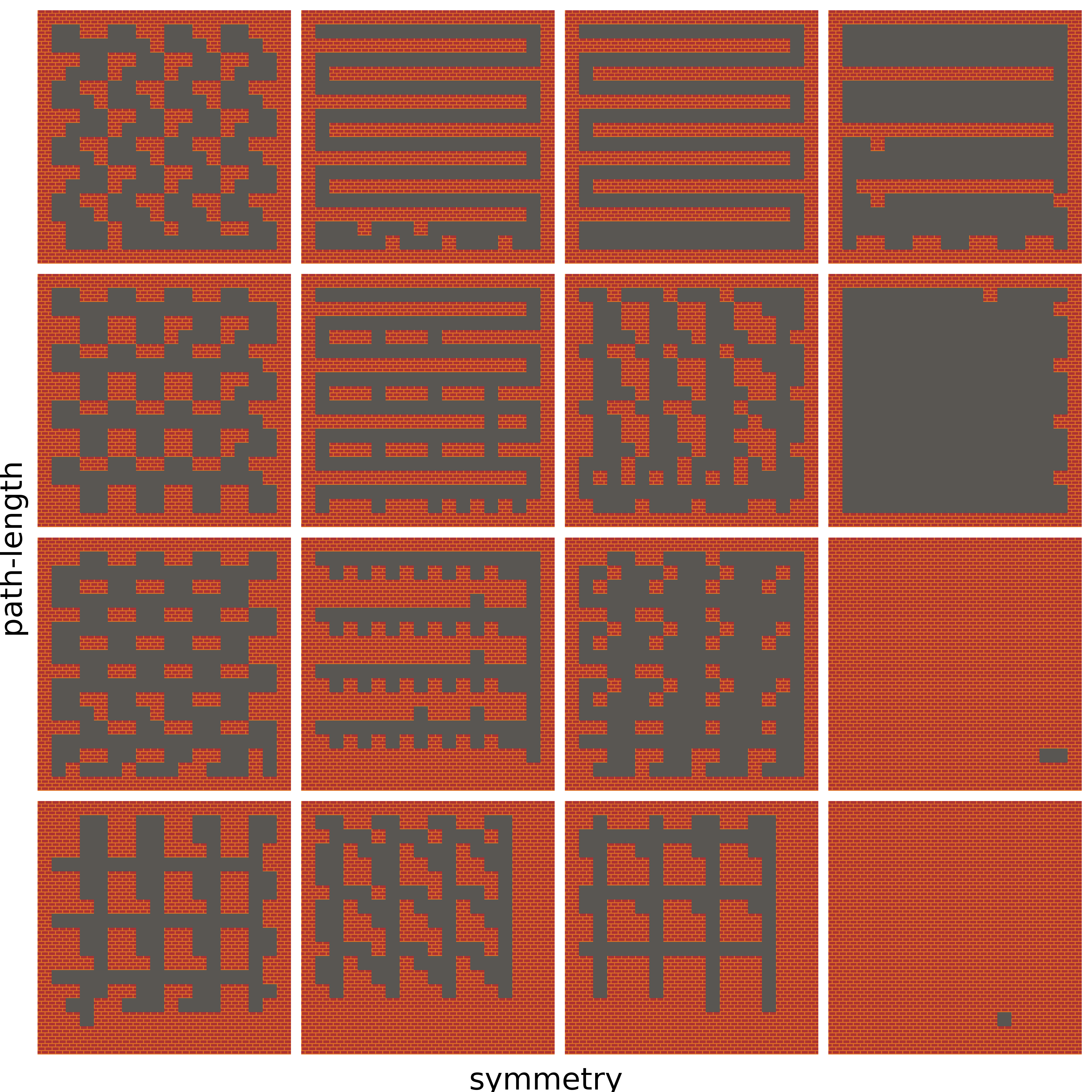}
      \includegraphics[width=.49\linewidth,trim=0 0 20 0,clip]{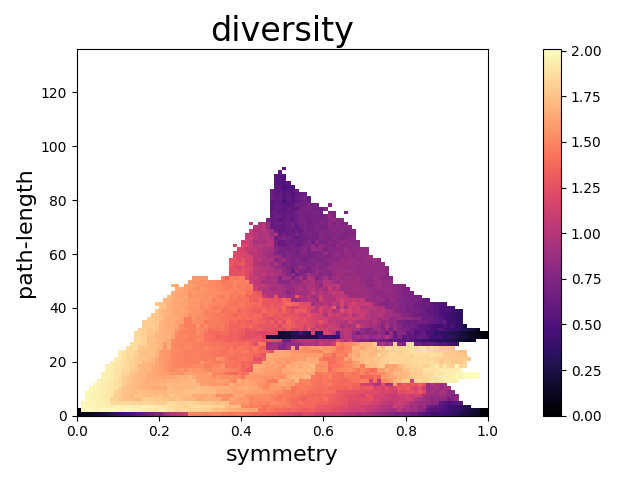}
      \includegraphics[width=.49\linewidth,trim=0 0 30 0,clip]{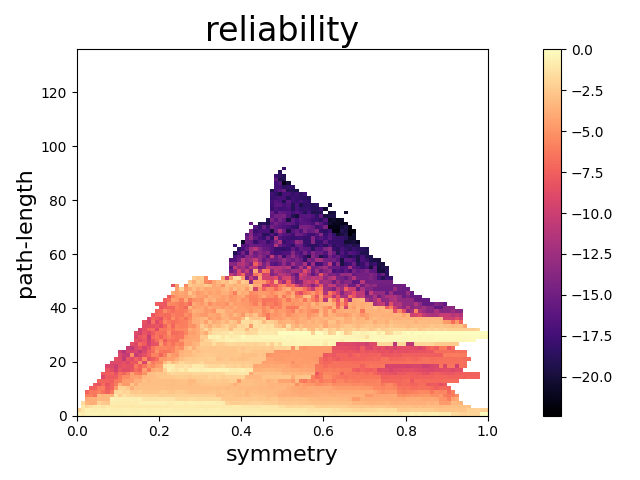}
      \caption{\textbf{Decoder} generators.}
      \label{fig:binary_cppn}
    \end{footnotesize}
  \end{subfigure}
  \begin{subfigure}[t]{0.49\textwidth}
    \begin{footnotesize}
      \centering
      \includegraphics[width=1.0\textwidth]{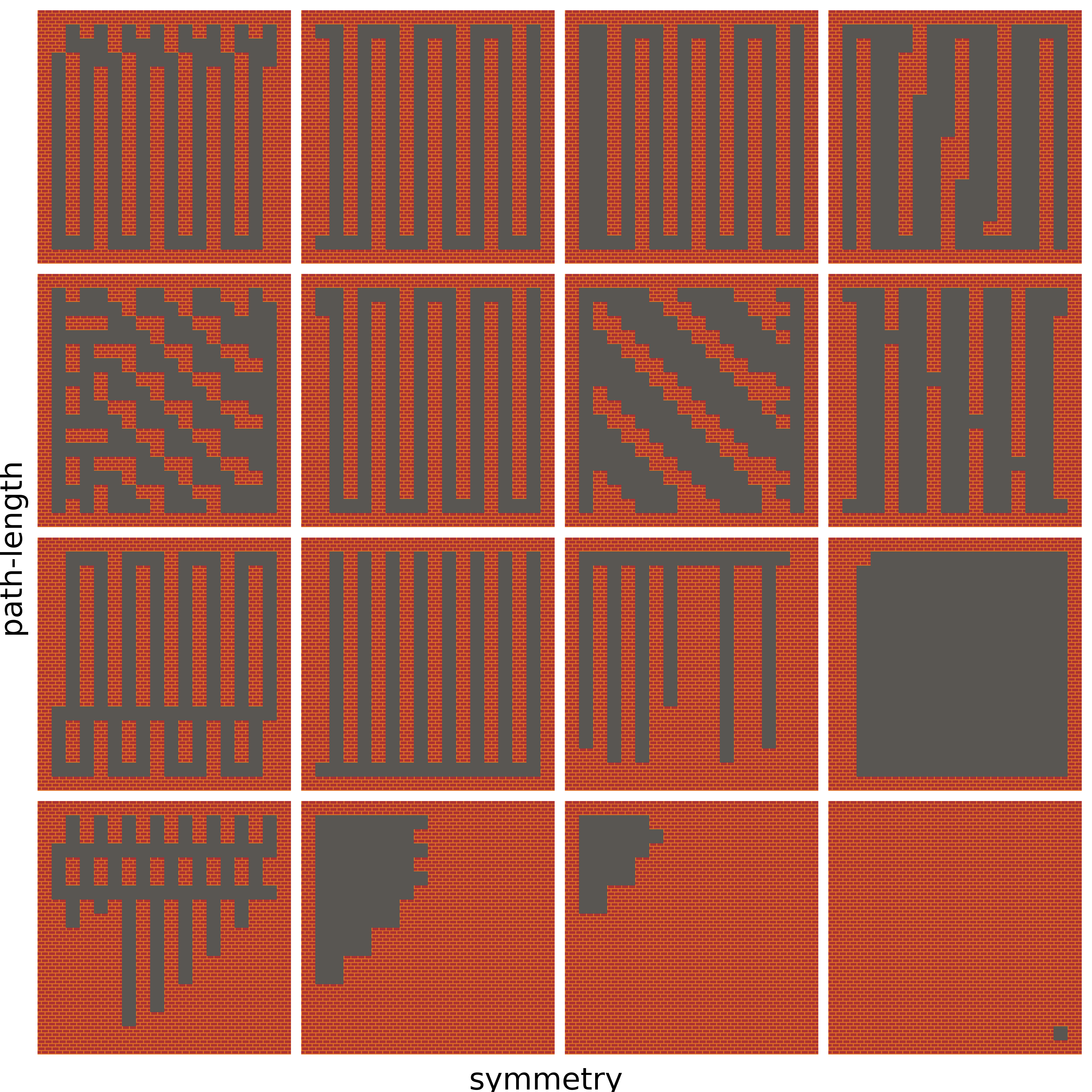}
      \includegraphics[width=.49\linewidth,trim=0 0 30 0,clip]{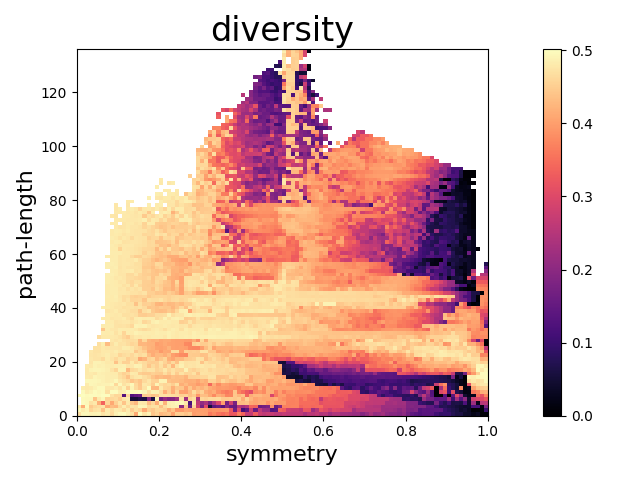}
      \includegraphics[width=0.49\linewidth,trim=0 0 30 0,clip]{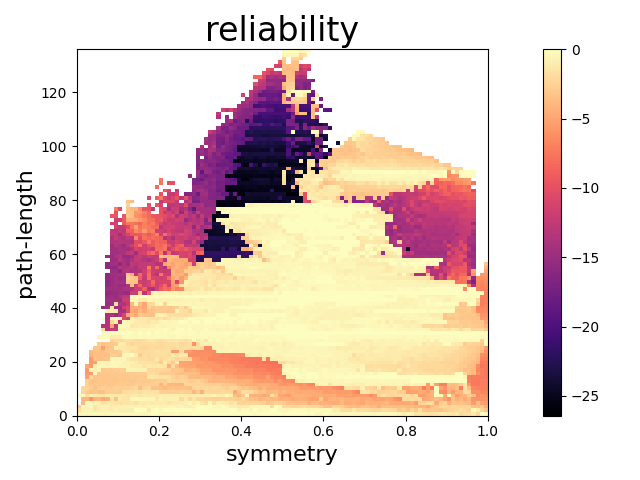}
      \caption{\textbf{NCA} generators.}
      \label{fig:binary_nca}
    \end{footnotesize}
  \end{subfigure}
  \caption{Decoder-style architectures use global information and map latent seeds to levels in a single pass, while NCAs produce global coherence by iterative local interactions. With symmetry and path-length as measures of interest, NCAs yield a more thorough exploration of level space, filling out more of the archive.}
  \label{fig:binary_levels}
\end{figure*}

\subsection{The objectives of level generators}

We define an objective function $o$ for a given generator as a weighted sum of components corresponding to validity $v$, reliability $r$, and intra-generator diversity $d$ of a batch of levels $L$ produced by the generator for a given game.

\subsubsection{Validity}\label{sec:validity_Only}

The validity term captures how well the batch of generated levels conform to soft constraints required by the game and each game has different validity constraints. For example, in Sokoban and Zelda a desired constraint is the level forming one connected region. To encourage this constraint we count the number of connected components present in the level and penalize generators for each additional region formed. In Zelda, valid levels contain $[2,5]$ enemies. We penalize how far outside the range a generated level falls. We give a full description of all validity constraints and penalties in the supplementary material.

Once we aggregate all validity penalties into a negative penalty term $v_l$ for a given level $l$ in batch $L$, we average all penalties to form the validity term $v$:

\begin{equation}\label{eq:validity}
\begin{split}
  v &= \frac{1}{|L|}\sum_{l\in L} v_l
\end{split}
\end{equation}



A perfectly valid level will obtain $v_l = 0$, while invalid levels will have a negative $v_l$ term.




\subsubsection{Reliability}

The reliability term captures the precision a generator maintains in measurement space. The batch of levels $L$ produced by the generator may have different outputs for the measure functions. For example, in the maze domain the length of the maze solution is a measure function for the QD problem. In this case, reliability captures how consistently a generator can produce levels of a given length.


For each measure $m_i \in M$, the reliability penalty $r_i$ is the standard deviation of the batch of levels $L$ along measure $m_i$. We then take the average reliability score over the set of measures to arrive at the generator's reliability penalty $r$:


\begin{equation}\label{eq:reliability}
r = - \frac{1}{|M|} \sum_i r_i
\end{equation}

\subsubsection{Diversity}

Intra-generator diversity refers to the amount of variation among the batch of levels produced by a generator. The term discourages generators from collapsing to a single optimal level, ignoring latent seed inputs.


The diversity bonus $d$ is the mean pairwise hamming distance over the batch of levels, normalized per-tile; or the frequency (as a ratio from 0--1) with which a given coordinate tile differs between any two levels:

\begin{equation}\label{eq:diversity}
d = \frac{1}{(h \cdot w \cdot |L|^2)-1} \sum_{i \in L} \sum_{j \in L} H(i,j)
\end{equation}

where \(w\) and $h$ are the width and height of the levels, respectively, and \(H(i,j)\) is the hamming distance between levels \(i\) and \(j\) (viz. the number of cells on the level-grid that are not occupied by the same type of tile on both levels).

\subsubsection{Combining Terms}\label{sec:combo_obj}
Finally, we combine all three terms into a weighted sum to form the objective function $o$:

\begin{equation}\label{eq:objective}
o = v + \text{max}(0, r + 10 d)
\end{equation}

To understand the weighting of the terms, first consider how reliability and diversity affect one another. The desired goal of a generator is to produce a distribution of valid levels with precise measure values. One issue is a generator can obtain trivial diversity by having low reliability. So unreliability should detract from the diversity bonus, but not the validity score. Note that the output range of reliability is approximately $[-50, 0]$ and the diversity output range is a ratio in the range $[0, 1]$. We weight diversity to be closer to the range of reliability and then take the maximum with 0. The term $\max(0, r + 10 d)$ ensures that the reliability penalty only contributes when diversity is high enough. Finally, we add the $\max(0, r + 10 d)$ term to the validity penalty term to form our final objective function to be maximized in the QD problem.

\subsection{Game environments}

We generate collections of NCA generators for several game environments.
In the maze domain, we aim to construct mazes with variable path-length (i.e. the longest shortest path between any two empty tiles), and in which all empty tiles are reachable by any other.
In Zelda, levels must contain one player, key, door, 2--5 enemies, and again, strictly reachable empty tiles. The length of the path from player to key to door should vary, and the nearest enemy should be at least 4 tiles away from the player.
In Sokoban, where the player must push each crate onto a target in order to complete the level, there must be the same number (and at least one each) of crates and targets, and the length of the solution produced by a deterministic solver should be varied.

\section{Results and discussion}

\begin{figure}[ht]
  \begin{footnotesize}
    \centering
    \includegraphics[width=.495\textwidth]{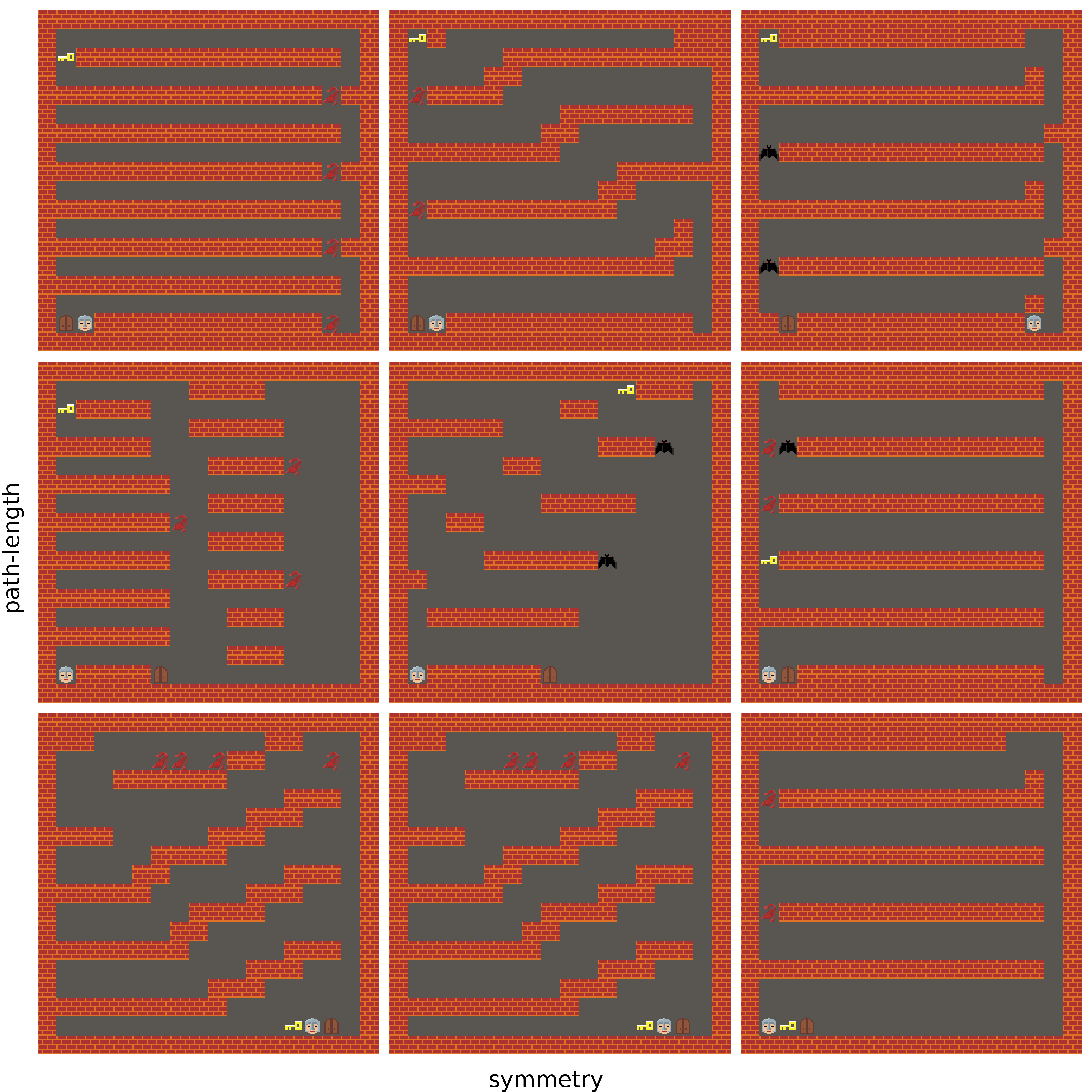}
    \includegraphics[width=.49\linewidth,trim=0 0 30 0,clip]{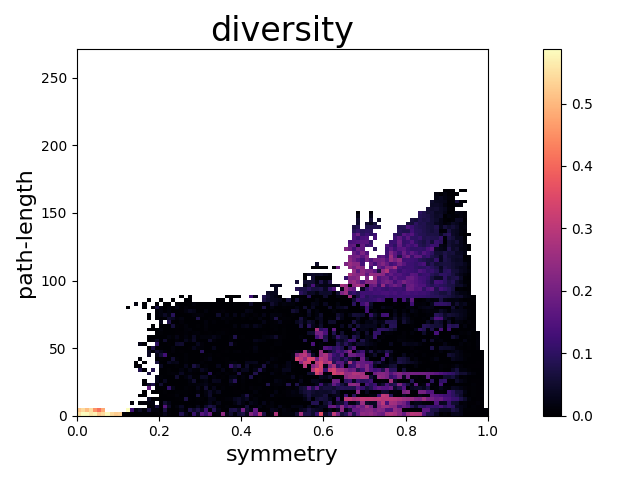}
    \includegraphics[width=.49\linewidth,trim=0 0 30 0,clip]{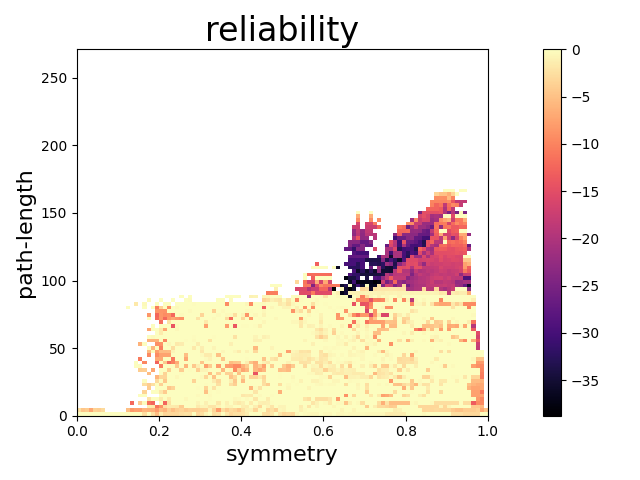}
    \caption{\textbf{Zelda levels produced by NCA generators diverse along measures of symmetry and path-length.} The levels with the longest paths tend to be more symmetrical.}
    \label{fig:zelda_levels}
  \end{footnotesize}
\end{figure}

\begin{figure}[ht]
  \begin{footnotesize}
    \centering
    \includegraphics[width=.495\textwidth,]{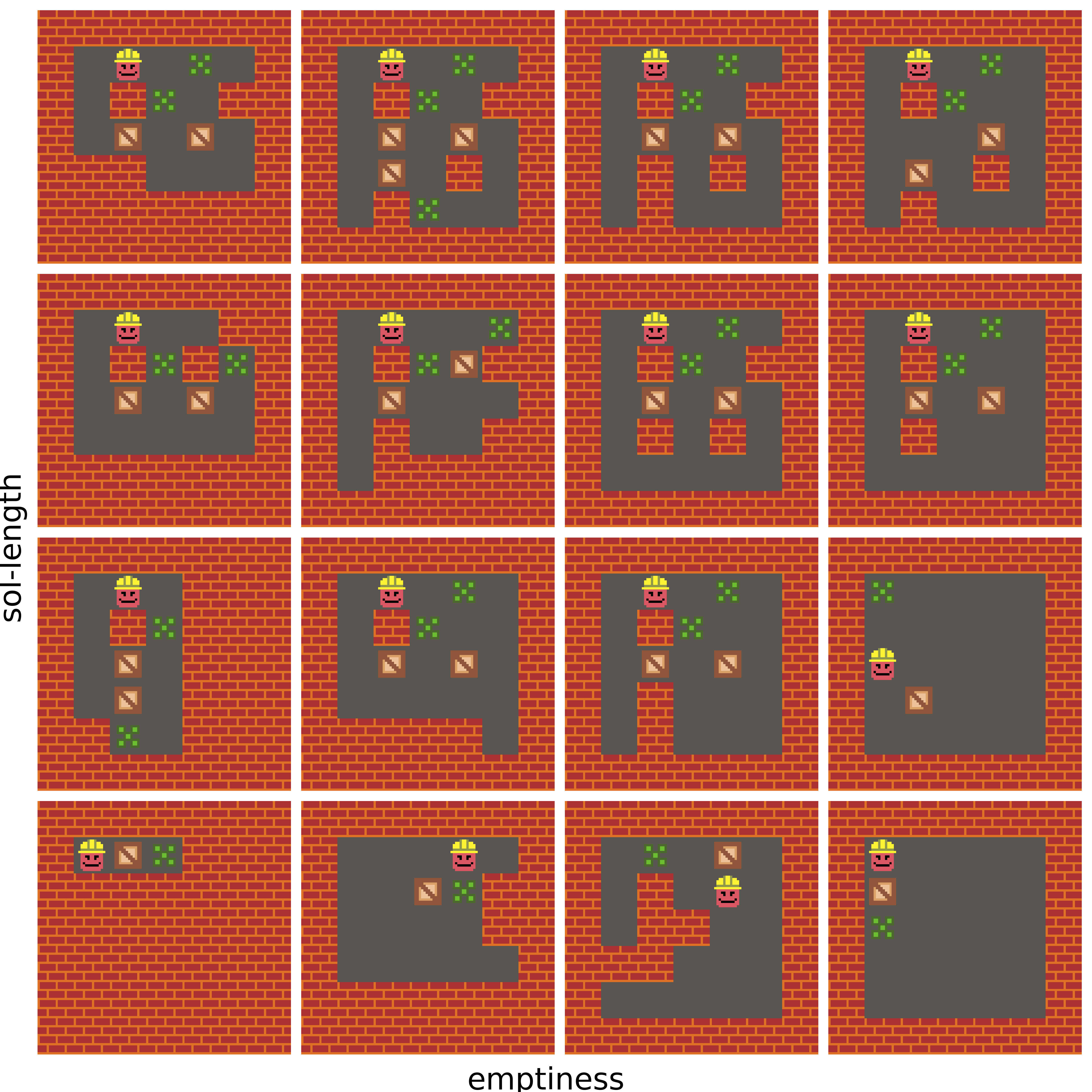}
    \includegraphics[width=.495\linewidth,trim=0 0 30 0,clip]{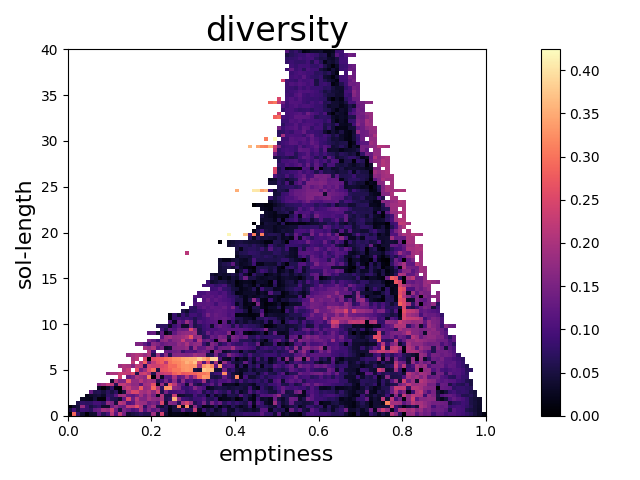}
    \includegraphics[width=.495\linewidth,trim=0 0 30 0,clip]{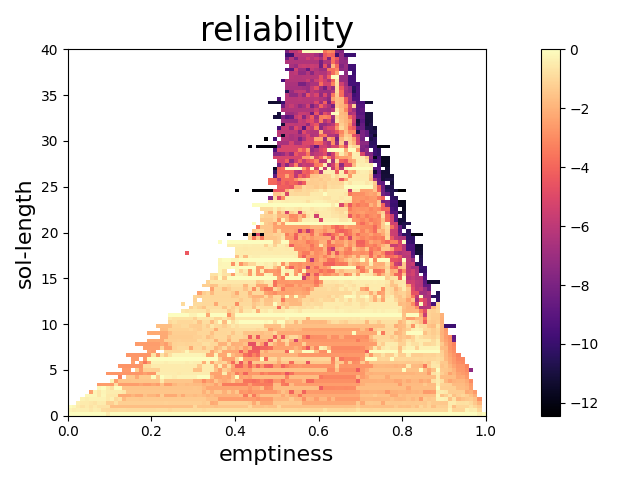}
    \caption{\textbf{Sokoban levels, diverse along measures of emptiness and path-length.} Levels requiring the most lengthy solutions tend to be roughly half-empty, with obstacles arranged so as to demand significant back-tracking from the player.}
    \label{fig:sokoban_levels}
  \end{footnotesize}
\end{figure}

All experiments are run on a single 48--CPU node.
The experiments in Tables~\ref{tbl:cross_eval_models} and \ref{tbl:cross_eval} and Figures~\ref{fig:binary_levels}--\ref{fig:sokoban_levels} were run for $50,000$ iterations each, taking 3 days or less. In Tables~\ref{tbl:cross_eval_models} and~\ref{tbl:cross_eval}, each experiment is repeated 11 times, and we report means and standard deviations over these trials. 

\begin{table*}
\centering
\caption{Neural architecture performance on maze-generation task, with exploration along measures of path-length and symetry. NCAs outperform Decoders, as well as indirect-encoding and generative, mutable and fixed-topology CPPNs, in terms of archive size and Quality Diversity score during training, and maintain the highest QD score during evaluation. Intra-generator diversity is significantly higher among NCAs during both training and evaluation. Note that non-generative (indirect-encoding) CPPNs are not evaluated on new latent seeds, since they take only $(x, y)$ coordinates as input.}
\label{tbl:cross_eval_models}
\label{{'tbl:cross_eval'}}
\begin{tabular}{|r|r|r|r|r|r|r|r|r|r|r|r|r|r|r|r|r|r|r|r|r|r|r|r|r|r|r|r|r|r|r|r|r|r|r|r|r|r|r|r|r|r|r|r|r|r|r|r|}
\hline
         &    & \multicolumn{3}{c|}{Training} & \multicolumn{3}{c|}{Evaluation} \\
\cline{3-8}
         &    &            archive size &             QD score & \begin{tabular}[c]{@{}l@{}}generator\\ diversity\end{tabular} &            archive size &             QD score & \begin{tabular}[c]{@{}l@{}}generator\\ diversity\end{tabular} \\
\textbf{model} & \textbf{\begin{tabular}[c]{@{}l@{}}num.\\ steps\end{tabular}} &                         &                      &                                                               &                         &                      &                                                               \\
\hline
\textbf{CPPN} & \textbf{1 } &           4,959 ± 893.1 &           424 ± 35.2 &                                          --- &  --- &           --- &                                         --- \\
\textbf{SinCPPN} & \textbf{1 } &            1,930 ± 13.1 &            233 ± 1.6 &                                          --- &            --- &            --- &                                         --- \\
\textbf{Decoder} & \textbf{1 } &           3,502 ± 167.8 &           429 ± 21.2 &                                \textbf{1.14} ± 0.1 &              828 ± 52.6 &             94 ± 6.4 &                                \textbf{1.29} ± 0.0 \\
\textbf{GenCPPN} & \textbf{1 } &           2,003 ± 142.6 &           250 ± 17.2 &                                         0.32 ± 0.0 &            1,078 ± 61.1 &            134 ± 7.5 &                                         0.35 ± 0.0 \\
\textbf{GenSinCPPN} & \textbf{1 } &               619 ± 6.4 &             71 ± 0.7 &                                         0.33 ± 0.0 &               220 ± 6.5 &             25 ± 0.7 &                                         0.33 ± 0.0 \\
\textbf{NCA} & \textbf{50} &           4,584 ± 456.2 &           860 ± 57.1 &                                         0.37 ± 0.0 &           3,122 ± 289.9 &           386 ± 36.6 &                                         0.36 ± 0.0 \\
\textbf{AuxNCA} & \textbf{50} &  \textbf{7,118} ± 802.0 &  \textbf{910} ± 95.0 &                                         0.23 ± 0.1 &           \textbf{4,035} ± 596.9 &  \textbf{501} ± 76.3 &                                         0.23 ± 0.1 \\
\hline
\end{tabular}
\end{table*}

\subsection{Comparison of architectures}

In Table \ref{tbl:cross_eval} (appendix), we present the results of several hyperparameter ablations. Given the vanilla NCA architecture, we evolve archives evaluated on 10 or 20 latent seeds or random initial level states (denoted as ``batch size''), either fixing or resampling these latents throughout evolution, and given level-generation episodes of either 50 or 100 iterations.

In Table \ref{tbl:cross_eval_models}, we focus on the effect of model architectures. We fix other hyperparameters, using a batch size of 10, fixed latent seeds, and 50-step episodes in the case of NCAs.

In both cases, we examine the performance of evolved archives of neural network-based level-generators, both on the same latent seeds that were seen during training, and on 20 new seeds sampled during evaluation. After training, we observe the size of each archive as well as its quality diversity score (the sum of all generators' non-negative objective score\footnote{Each elite's objective score from Equation \ref{eq:objective} minus the minimum possible objective/validity score})~\citep{pugh2015confronting}, and the mean intra-generator diversity score (\(d\), Equation \ref{eq:diversity}). 

We note that the QD scores are lower in evaluation, where we use new seeds, compared to using the same seeds as training. This is because during evaluation, generators are prompted with new latent inputs and re-inserted into a fresh archive. Typically, this causes archives that are adversely affected by new latents to produce less diverse measure scores. Elites fall back toward the center of the archive, where they often collide in the same cell, causing a reduction in the size of the archive and the archive's QD score.

In Table~\ref{tbl:cross_eval_models}, we see that NCA models maintain the largest archive and highest QD score after evaluation on new latent seeds, suggesting that they are more robust generators when compared against other architectures. Generative CPPNs (\textbf{GenCPPN}s) perform next-best in these metrics during evaluation, followed by the \textbf{Decoder}, and finally fixed-topology sinusoidal CPPNs (\textbf{GenSinCPPN}s).

\textbf{AuxNCA}s---NCAs with additional auxiliary channels---outperform vanilla \textbf{NCA}s. On the other hand, the former exhibit less intra-generator diversity, while vanilla NCAs are on par with Generative CPPNs in this regard. Although the Decoder architecture has the highest intra-generator diversity, it performs quite poorly in other metrics, and we note that for any given generator, producing diverse output becomes trivial as the generated levels decrease in quality and fail to meet the various constraints being imposed upon them. 



During training, NCAs result in archives with the highest QD scores. They also result in the largest archives among generative architectures. NCAs with auxiliary channels produce the largest archives among all methods, while vanilla NCAs result in archives whose size is roughly on par with those populated by (non-generative, indirect-encoding) CPPNs. This suggests that---even putting their generative abilities aside---NCA-based architectures are better suited for searching the space of high-quality and diverse levels (given some fixed set of latent seeds).

The differences in archive size, QD score, and intra-generator diversity across model architectures during training and evaluation was statistically significant as determined by one-way ANOVA tests ($p < 0.001$). Tukey post hoc tests additionally revealed that the difference in each of these metrics at training and evaluation time between each pair of models was statistically significant ($p = 0.001$) in nearly all cases. In all cases, significance holds between the dominant model and all others in a given metric. Both NCA models significantly outperform all others in terms of both archive size and QD score at training and evaluation time, with the only exception being between the archive sizes resulting from the vanilla NCAs and the non-generative CPPNs during training.

\begin{table*}
\centering
\caption{\textbf{Hyperparameter ablations in the maze domain on the NCA model, with exploration along measures of path-length and symmetry.}}
\label{tbl:cross_eval}
\begin{tabular}{|r|r|r|r|r|r|r|r|r|r|r|r|r|r|r|r|r|r|r|r|r|r|r|r|r|r|r|r|r|r|r|r|r|r|r|r|r|r|r|r|r|r|r|r|r|r|r|r|r|r|}
\hline
    &    &           &     & \multicolumn{3}{c|}{Training} & \multicolumn{3}{c|}{Evaluation} \\
\cline{5-10}
    &    &           &     &            archive size &             QD score & \begin{tabular}[c]{@{}l@{}}generator\\ diversity\end{tabular} &            archive size &             QD score & \begin{tabular}[c]{@{}l@{}}generator\\ diversity\end{tabular} \\
\textbf{model} & \textbf{\begin{tabular}[c]{@{}l@{}}batch\\ size\end{tabular}} & \textbf{latents} & \textbf{\begin{tabular}[c]{@{}l@{}}num.\\ steps\end{tabular}} &                         &                      &                                                               &                         &                      &                                                               \\
\hline
\multirow{8}{*}{\textbf{NCA}} & \multirow{4}{*}{\textbf{10}} & \multirow{2}{*}{\textbf{Fix}} & \textbf{50 } &           4,584 ± 456.2 &           860 ± 57.1 &                                         0.37 ± 0.0 &           3,122 ± 289.9 &           386 ± 36.6 &                                         0.36 ± 0.0 \\
    &    &           & \textbf{100} &           6,804 ± 342.8 &           849 ± 42.4 &                                         0.37 ± 0.0 &           3,162 ± 286.9 &           392 ± 35.5 &                                         0.37 ± 0.0 \\
\cline{3-10}
    &    & \multirow{2}{*}{\textbf{Re-sample}} & \textbf{50 } &           6,442 ± 359.1 &           797 ± 42.7 &                                         0.35 ± 0.0 &           3,396 ± 207.6 &           417 ± 24.4 &                                         0.34 ± 0.0 \\
    &    &           & \textbf{100} &           6,560 ± 295.5 &           816 ± 38.3 &                                         0.35 ± 0.0 &           3,477 ± 186.3 &           430 ± 24.0 &                                         0.35 ± 0.0 \\
\cline{2-10}
\cline{3-10}
    & \multirow{4}{*}{\textbf{20}} & \multirow{2}{*}{\textbf{Fix}} & \textbf{50 } &           6,650 ± 491.1 &           826 ± 62.4 &                                         0.35 ± 0.0 &           3,458 ± 454.2 &           428 ± 56.7 &                                         0.35 ± 0.0 \\
    &    &           & \textbf{100} &  \textbf{6,938} ± 365.4 &  \textbf{863} ± 48.3 &                                         0.37 ± 0.0 &           3,596 ± 278.8 &           446 ± 33.8 &                                         0.35 ± 0.0 \\
\cline{3-10}
    &    & \multirow{2}{*}{\textbf{Re-sample}} & \textbf{50 } &           6,458 ± 618.7 &           798 ± 82.0 &                                         0.34 ± 0.0 &           3,683 ± 385.7 &           454 ± 50.9 &                                         0.34 ± 0.0 \\
    &    &           & \textbf{100} &           6,604 ± 269.8 &           817 ± 36.3 &                                \textbf{0.37} ± 0.0 &  \textbf{3,747} ± 207.4 &  \textbf{462} ± 28.1 &                                \textbf{0.37} ± 0.0 \\
\hline
\end{tabular}
\end{table*}

In Table~\ref{tbl:cross_eval}, we find that our experiments are robust to changes in these hyperparameters, with no statistically significant differences resulting from hyperparameter changes other than the model architecture itself.

\subsection{Level quality \& diversity}


In the maze domain, an archive of generators is optimized along measures of symmetry (horizontal and vertical) and path-length in the mazes they produce (Figure~\ref{fig:binary_levels}).
After 50,000 generations, both \textbf{NCA} and \textbf{Decoder} architecutures, evolved with CMA-ME have filled out a considerable portion of the archive, and come close to maximum path-length zig-zags. These near-optimal levels tend to occur in a region of level-space with medium symmetry.
But NCAs fill out more of the high-path-length extremeties of the archive, where generators tend to be less reliable along measures of interest, and often less diverse.

These results suggest that generators building levels in an emergent manner through iterative local interactions are able to explore level-space more thoroughly than one-step generators with global information.

Where Decoder and NCA archives overlap, NCAs produce levels with target metrics of interest more reliably, although limiting diversity in the process.

In Zelda (Figure~\ref{fig:zelda_levels}), we search along measures of symmetry and path-length. The maximum path-length level is further out of reach of the archive of NCAs. However, CMA-ME has explored along an area of level-space with high symmetry, culminating in something like a two-lane version of the optimal zig-zag (top-right).

In Sokoban (Figure \ref{fig:sokoban_levels}), an archive of generators is optimized to explore along measures of emptiness and solution length. Optimization first discovers small, valid corridors, then extends them to produce chambers of increasing complexity.
Again, more complex generators tend to be more brittle.

Intra-generator diversity among Zelda and Sokoban level-generators is considerably lower than among maze generators. This may be because designing Zelda levels involves more interrelated constraints than when designing mazes: moving a player, door or key will not just potentially affect symmetry, but also path-length. The reliability score may also be weighted too heavily.

\section{Limitations and future work}

In this work, the generators are optimized to produce levels with exactly the same values along measures of interest (Equation~\ref{eq:reliability}). This ensures that the resulting archive can produce levels whose features are controllable with high precision.
But for the sake of the optimization process, it may also be desirable to treat this variance as a QD measure.
Depending on the end-user's need for reliability, they could then select only sufficiently reliable generators from the final archive during actual level design.
By instead treating reliability as an objective, we may be stifling search from discovering bridges to high-complexity areas of level space via unreliable but exploratory generators.

As this work argues for the viability of NCAs as level generators, a direct comparison against other neural networks that have already been used to this end would be desirable.
But the size of these latter networks makes covariance matrix adaptation of their weights computationally infeasible for methods like CMA-ME, as this matrix has size $n^2$, with $n$ equal to the number of parameters in the network.
Future work might side-step this issue by using differentiable quality diversity (DQD)~\citep{fontaine2021differentiable}, which, given a differentiable genome, uses gradients to update a covariance matrix that instead only scales quadratically with the number of diversity measures.

While the low parameter-count of NCAs is integral to the feasability of our approach, each NCA update only propagates local changes, leading to many steps needed to generate each level and the use of the intermediate levels as external memory.
These dependencies, on episode length and the number of (one-hot) tile-types on the game-board, could in theory be foregone by applying the NCA's single $3\times 3$ convolutional layer repeatedly.
In a similar vein, fractal neural networks with weight-sharing \citep{larsson2016fractalnet, earle2020using} incorporate stabilizing, structured skip connections, either without any new weights, or with new weights added logarithmically with respect to increases in the size of the model's perceptive field.


In our experiments, we have compared NCAs with several neural network-based generator representations that were trained with versions of MAP-Elites. There are many other methods for generating 2D levels that we have not compared with, because they fundamentally solve a different problem, so that a fair comparison would not be possible. For example, we could have used MAP-Elites to evolve levels using a direct representation~\cite{alvarez2019empowering}, or in the latent space of a generator network~\cite{fontaine2021illuminating}, but this would have meant searching for levels instead of level generators. Conversely, one could learn level generators using for example Generative Adversarial Networks or WaveFunctionCollapse, but this would have required example levels to train on, and would not generate archives of multiple generators that differ along pre-specified measures. In fact, we are not aware of previous attempts at generating archives of \emph{generators} using QD algorithms.

\section{Conclusion}

We present a method for generating a diverse archive of level-generators represented as simple Neural Cellular Automata. The space of NCAs for growing levels is explored by CMA-ME along measures of interest (like symmetry or solution-length) in the levels they produce.
Each generator in the archive is optimized to produce batches of valid levels with reliable features along these measures, with these batches of levels varying  among themselves where possible.

While NCAs are restricted to local computation, they propagate spatial information across the level while editing it sequentially to form patterns with global complexity.
These NCAs are capable of satisfying high-level functional constraints and heuristics in mazes, and zelda and sokoban levels (producing, e.g. puzzles that require a certain number of steps for completion by an A* solver).


Compared to models with global information (fixed-topology and mutable CPPNs, optimized with CMA-ME and MAP-Elites, respectively; and Decoders optimized with CMA-ME), NCAs facilitate a more thorough exploration of level space, and its individual generators have much more diverse outputs.

The archive of level-generators produced by our method have potential as co-creative tools, in which the human user might select from a archive of generators with diverse aesthetic and functional tendencies, perturbing their design process to dynamically explore the space of playable levels. They could also work for Experience-Driven Procedural Content Generation, where generators are chosen from an archive to produce a target experience for a particular induced player profile~\cite{yannakakis2011experience}.
They may also be used to generate automatic curricula of game levels for training player-agents.
To this end, our method explicitly incentivizes both the reliability of the generator along measures of interest, and the generation of diverse levels at each point in level-space.
This could help increase the precision of the generated curriculum, and prevent a learning agent from over-fitting to any one difficult portion of the curriculum in which it would be fated to spend much of its time.

\bibliographystyle{ACM-Reference-Format}
\bibliography{biblography}


\end{document}